\pdfoutput=1
\documentclass[11pt]{article}
\usepackage{acl}
\usepackage[T1]{fontenc}
\usepackage[utf8]{inputenc}
\usepackage{times}
\usepackage{latexsym}
\usepackage{microtype}
\usepackage{inconsolata}
\usepackage{amsmath,amssymb,bm}
\usepackage{booktabs,multirow,makecell}
\usepackage{graphicx}
\usepackage{float}
\usepackage{tabularx}
\usepackage[table]{xcolor}
\usepackage{xparse}
\usepackage{threeparttable}
\usepackage[most]{tcolorbox}
\usepackage{caption} 
\usepackage{fontawesome}

\usepackage{hyperref} 

\definecolor{ClosedColor}{HTML}{D8EEF2}
\definecolor{OpenColor}{HTML}{FDEBDD}
\definecolor{ProprietaryColor}{HTML}{E3F2FD}   
\definecolor{Size400PlusColor}{HTML}{FFEBEE}   
\definecolor{Size100To400Color}{HTML}{FFF3E0}  
\definecolor{Size30To100Color}{HTML}{F9FBE7}   
\definecolor{Size10To30Color}{HTML}{E8F5E8}    
\definecolor{SizeBelow10Color}{HTML}{F3E5F5}   

\newcommand{\framework}{M$^{2}$G-Coder}
\newcommand{\benchmark}{M$^{2}$G-Eval}
\newcommand{\instruction}{\benchmark{}-Instruct}
\newcommand{\codersft}{\benchmark{}-Coder-SFT}
\newcommand{\coderl}{\benchmark{}-Coder-RL}
\newcommand{\languagenums}{18}

\title{\benchmark{}: Enhancing and Evaluating Multi-granularity Multilingual Code Generation}

\author{
  Fanglin Xu\textsuperscript{\rm 1},
  {\bf Wei Zhang}\textsuperscript{\rm 1 \thanks{\ Equal contribution. }},
  {\bf Jian Yang}\textsuperscript{\rm 1}\thanks{\ Corresponding author.},
  {\bf Guo Chen}\textsuperscript{\rm 2},
  {\bf Aishan Liu}\textsuperscript{\rm 1},
  {\bf Zhoujun Li}\textsuperscript{\rm 1}, \\
  {\bf Xianglong Liu}\textsuperscript{\rm 1} 
  {\bf Bryan Dai}\textsuperscript{\rm 3}\\
   \textsuperscript{\rm 1}Beihang University; 
   \textsuperscript{\rm 2}Hunan University;
   \textsuperscript{\rm 3}Ubiquant; \\
   \texttt{\{jiayang\}@buaa.edu.cn}\\
}
\begin{document}
\maketitle

\begin{abstract}
The rapid advancement of code large language models (LLMs) has sparked significant research interest in systematically evaluating their code generation capabilities, yet existing benchmarks predominantly assess models at a single structural granularity and focus on limited programming languages, obscuring fine-grained capability variations across different code scopes and multilingual scenarios. We introduce \benchmark{}, a multi-granularity, multilingual framework for evaluating code generation in large language models (LLMs) across four levels: Class, Function, Block, and Line. Spanning \languagenums{} programming languages, \benchmark{} includes 17K+ training tasks and 1,286 human-annotated, contamination-controlled test instances. We develop \benchmark{}-Coder models by training Qwen3-8B with supervised fine-tuning and Group Relative Policy Optimization. Evaluating 30 models (28 state-of-the-art LLMs plus our two \benchmark{}-Coder variants) reveals three main findings: (1) an apparent difficulty hierarchy, with Line-level tasks easiest and Class-level most challenging; (2) widening performance gaps between full- and partial-granularity languages as task complexity increases; and (3) strong cross-language correlations, suggesting that models learn transferable programming concepts. \benchmark{} enables fine-grained diagnosis of code generation capabilities and highlights persistent challenges in synthesizing complex, long-form code.
\end{abstract}

\section{Introduction}

The emergence of large language models (LLMs) specialized for code has fundamentally transformed software engineering practices. Modern code LLMs~\cite{starcoder,starcoder2,seedcoder,deepseek_coder}, such as KAT-Coder~\cite{katcoder} and Qwen3-Coder~\cite{qwen25coder}, leverage pre-training on massive code corpora to achieve remarkable performance across diverse programming tasks. These models power intelligent development environments, automate routine coding tasks, and assist developers in navigating complex codebases, thereby significantly accelerating software development cycles.

\begin{figure}[t]
\centering
\includegraphics[width=1.0\columnwidth]{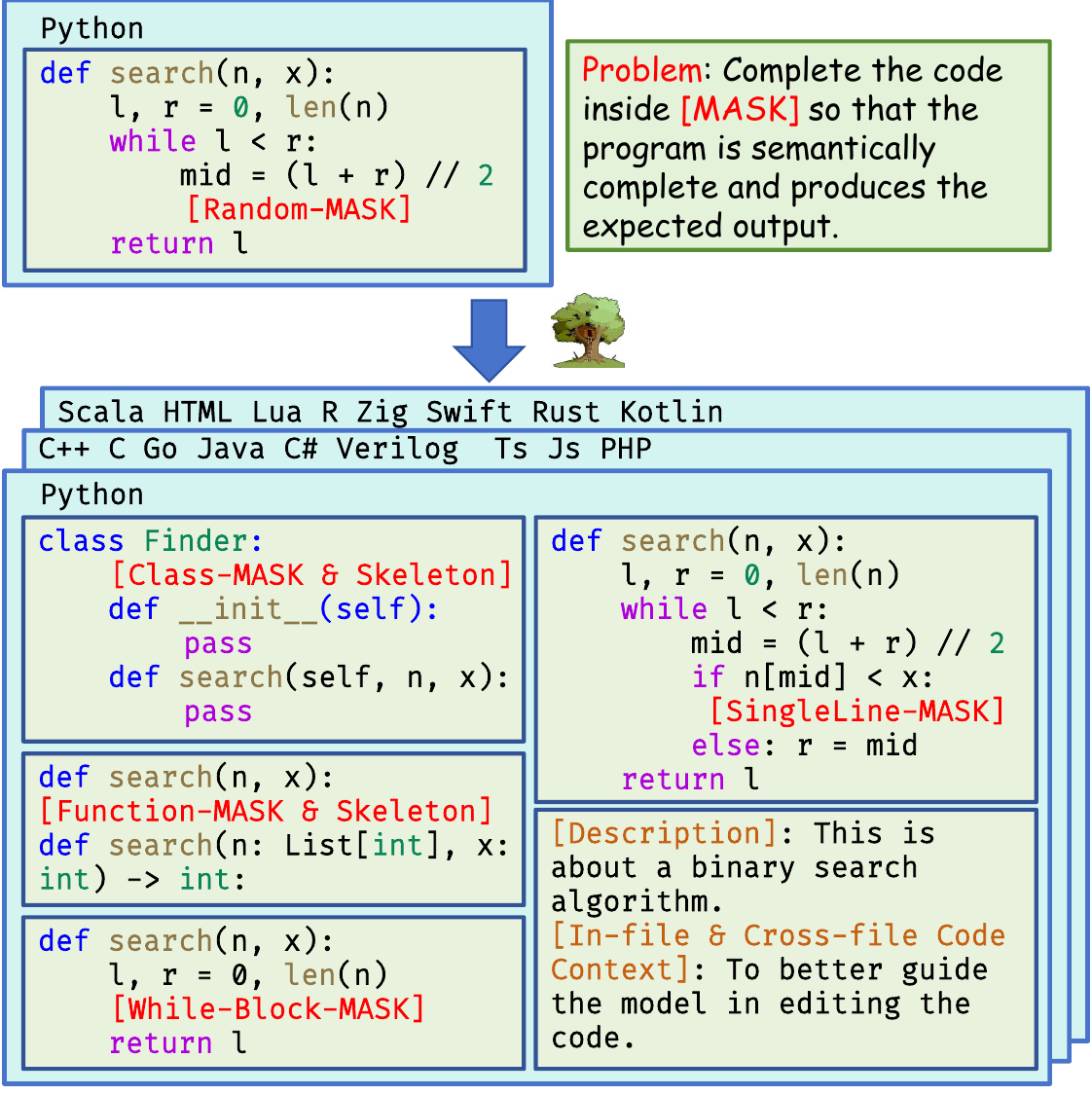}
\caption{\benchmark{} provides more challenging, multi-granularity code generation across more programming languages than previous work.}
\vspace{-25pt}
\label{fig:intro}
\end{figure}

\begin{figure*}[tb]
    \centering
    \includegraphics[width=1.0\textwidth]{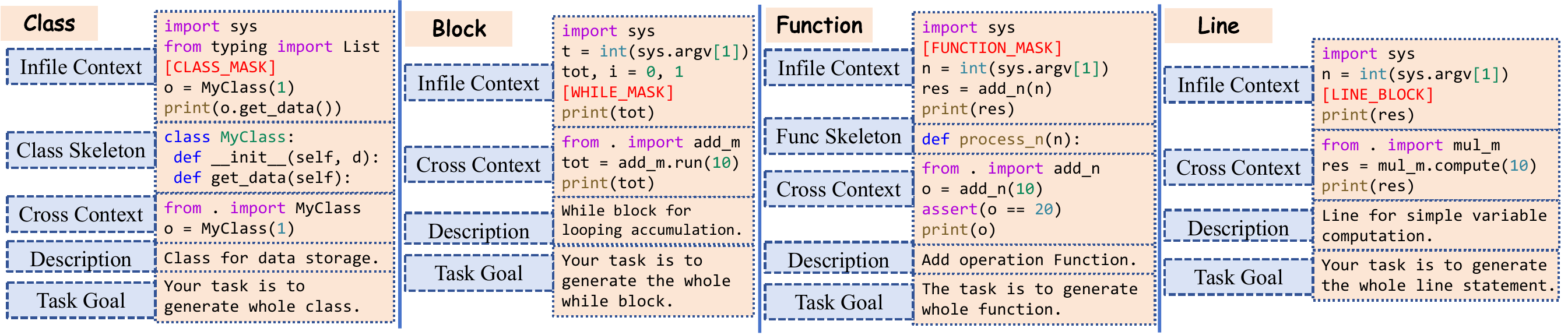}
    \caption{Four task granularity examples for \benchmark{}. Each example uses a simple Python code snippet to illustrate the data composition of Class, Function, Block, and Line-level tasks.}
    \vspace{-15pt}
    \label{fig:task-example}
\end{figure*}

Code generation represents a core capability of modern LLMs, underpinning applications from intelligent code completion to automated program synthesis. Early works focus on function-level code generation (e.g., HumanEval~\citep{codex} and MBPP~\citep{mbpp}), while recent works (e.g., CrossCodeEval~\citep{crosscodeeval}, M2RC-Eval~\citep{m2rceval}, and SWE-Bench~\cite{swebench}) assess repository-based capabilities. However, these frameworks adopt a \textit{single-granularity} evaluation paradigm, treating all code generation tasks uniformly, regardless of their structural scope. In reality, completing a single line of code requires fundamentally different contextual understanding and reasoning patterns than implementing a complete function or designing an entire class hierarchy. This granularity-agnostic approach obscures important variations in model capabilities across different code scopes. Furthermore, existing benchmarks predominantly focus on full-granularity languages such as Python and Java, with limited coverage of the diverse multilingual landscape characterizing real-world software ecosystems. \textit{Consequently, the community lacks a comprehensive evaluation framework that systematically measures code generation capabilities across multiple structural granularities and diverse programming languages.}

To address these limitations, we introduce \benchmark{}, a multi-granularity, multilingual framework that systematically enhances and evaluates code generation at four distinct structural levels: Class, Function, Block, and Line. In Figure~\ref{fig:intro}, \benchmark{} advances beyond existing benchmarks along two critical dimensions: (1) Finer-grained granularity, enabling differentiated assessment of model capabilities across code scopes, (2) Comprehensive language coverage, spanning \languagenums{} programming languages, including both full-granularity and partial-granularity languages. We first built \instruction{}, a large-scale instruction dataset containing about 17K training samples synthesized from roughly 150K repositories sampled from The-Stack-v2. Using abstract syntax tree parsing, we extract code units at multiple granularities and incorporate cross-file or in-file context for multilingual, multi-granularity supervised fine-tuning (SFT) and reinforcement learning (GRPO).
For evaluation, we construct \benchmark{} comprising 1,286 instances sourced from repositories created or updated after January 1, 2024, effectively mitigating pre-training data contamination. A team of 10 graduate and doctoral students with strong programming expertise manually validated each test instance, ensuring semantic accuracy, contextual completeness, and appropriate difficulty calibration.

The contributions are summarized as follows:
\begin{itemize}
    \item We introduce \benchmark{}, the first multi-granularity code-generation benchmark that systematically evaluates models across four structural levels (Class, Function, Block, Line) in \languagenums{} programming languages, featuring 1,286 human-annotated, contamination-controlled test instances.
    \item We construct \instruction{}, a large-scale instruction dataset with 17K+ high-quality training tasks derived from 150K repositories, employing Tree-Sitter-based parsing, BM25 cross-file retrieval, LLM-based description generation, and difficulty-calibrated filtering.
    \item We develop \benchmark{}-Coder models using a two-stage training pipeline (SFT followed by GRPO reinforcement learning) on Qwen3-8B, achieving strong performance and releasing both models to facilitate community research.
    \item We provide a comprehensive evaluation of 30 state-of-the-art LLMs, including two \framework{} models, revealing systematic patterns in granularity-dependent difficulty, language-resource disparities, and cross-lingual generalization, and establishing \benchmark{} as a rigorous diagnostic framework for assessing code-generation capabilities.
\end{itemize}

\begin{figure*}[htb]
    \centering
        \includegraphics[width=1.0\textwidth]{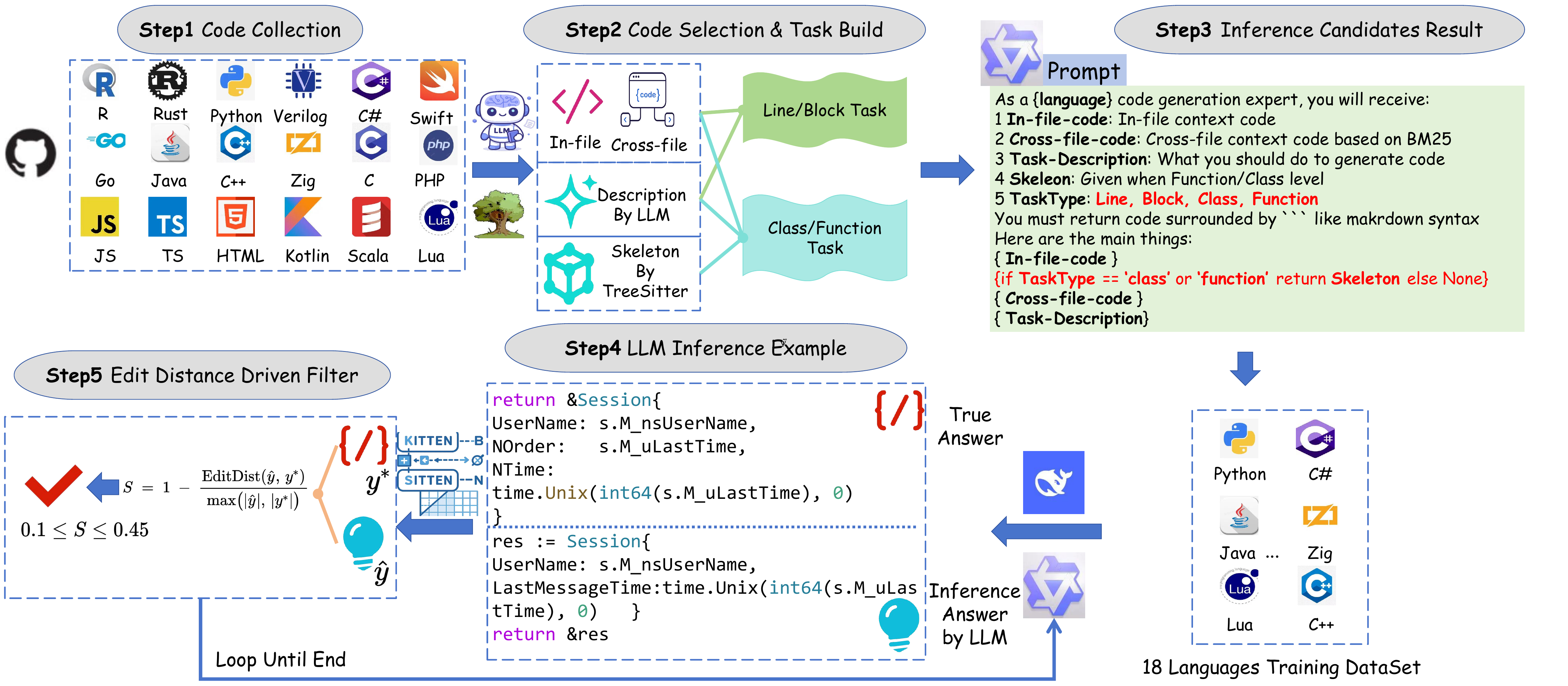}
    \caption{
    We construct \instruction{} by first curating sources across 18 languages, categorizing the materials, and instantiating four task granularities (class, function, block, line). Each task is wrapped as a structured prompt, after which we perform LLM-based quality filtering to obtain the final \instruction{}.}
    \vspace{-25pt}
    \label{fig:pipeline}
\end{figure*}

\section{Methodology} 

\subsection{\benchmark{} Task Definition}\label{sec:TaskDefinition}
\paragraph{Overall.}
We treat multi–granularity code generation as filling a masked region of code. Each example $\tau = (\ell, g, P, M, y^*)$ consists of a programming language $\ell$, a granularity label $g \in \{\text{Class}, \text{Function}, \text{Block}, \text{Line}\}$, a structured prompt $P$, a masked span $M$ aligned with $g$, and a reference implementation $y^*$. As illustrated in \autoref{fig:task-example}, the unified prompt
$P = (x_{\text{i}}, x_{\text{c}}, K, d, G)$
includes in-file context $x_{\text{i}}$, cross-file context $x_{\text{c}}$, an optional class or function skeleton $K$ (empty for Block and Line), an LLM-generated description $d$, and the task goal $G$. This provides a consistent input format for all four granularities. %

\paragraph{Inference Result.}
Models are required to return only the code that fills $M$, which we insert into $x_{\text{i}}$ to obtain the complete prediction $\hat{y}$. We then perform syntax and static checks, strip comments, normalize whitespace, and compute a length-normalized edit similarity $S \;=\; 1 \;-\; \frac{\mathrm{ED}(\hat{y},\,y^*)}{\max\!\big(|\hat{y}|,\,|y^*|\big)}
\label{eq:editdistance}$, where $\mathrm{ED}$ is the Levenshtein distance over token-id sequences from a fixed code tokenizer, and $|\cdot|$ is the token count. Higher $S$ indicates better agreement with the reference.

\subsection{\instruction{} Construction}\label{sec:TrainDataset}

\paragraph{Goal.} We construct the \instruction{} ($\mathcal{D}_{\text{t}}$) to train models for our multi-granularity task format. This instruction dataset serves both for supervised fine-tuning (SFT) and reinforcement learning (RL). To ensure quality, we apply a difficulty filter based on the edit similarity score $S$ to each candidate task.

\paragraph{Pipeline.} Our training dataset is built by the pipeline in \autoref{fig:pipeline}. We first sample about 150K repositories $R_{\text{t}}$ from The-Stack-v2~\citep{lozhkov2024starcoder} covering 18 languages and collect their source files. To reduce noise and boilerplate, we strip comments and configuration-heavy dependencies while preserving executable semantics. We then use Tree-Sitter\footnote{\url{https://tree-sitter.github.io/tree-sitter/}} to parse each file, locate editable units, and extract the in-file context $x_{\text{i}}$, the target code $y^*$, and the masked span $M$. Qwen3-Coder-480B-A35B-Instruct~\citep{qwen3technicalreport}, denoted $G_{\text{t}}$, generates a natural-language description $d$ for each snippet. For Class and Function tasks, we also extract the skeleton $K$ (e.g., a class's fields and methods, or a function's signature). To enrich context, we apply BM25 over the repository to retrieve related code as cross-file context $x_{\text{c}}$, yielding the initial dataset $\mathcal{D}'_{\text{t}}$. Finally, we run $G_{\text{t}}$ again to produce draft solutions $\hat{y}$, compute the similarity score $S$, and retain only tasks with $S$ between $0.1$ and $0.45$, resulting in the final training data $\mathcal{D}_{\text{t}}$ with about 17K tasks.

\subsection{\benchmark{} Dataset Construction}\label{sec:TestDataConstruct}%

\paragraph{Goal.} We constructed the training dataset $\mathcal{D}_{\text{t}}$ in \autoref{sec:TrainDataset}.
Building on this, we construct an independent, high-quality evaluation dataset, $\mathcal{D}_{\text{e}}$, to rigorously evaluate the performance of \codersft{} and \coderl{} and to ensure a fair comparison with other baseline models. The core goals of this dataset are authoritativeness and being free from pretraining data contamination. We ensure that the evaluation dataset is disjoint from the training data, such that $\mathcal{D}_{\text{e}} \cap \mathcal{D}_{\text{t}} = \varnothing$.

\begin{figure*}[tb]
\centering
    \includegraphics[width=1.0\textwidth]{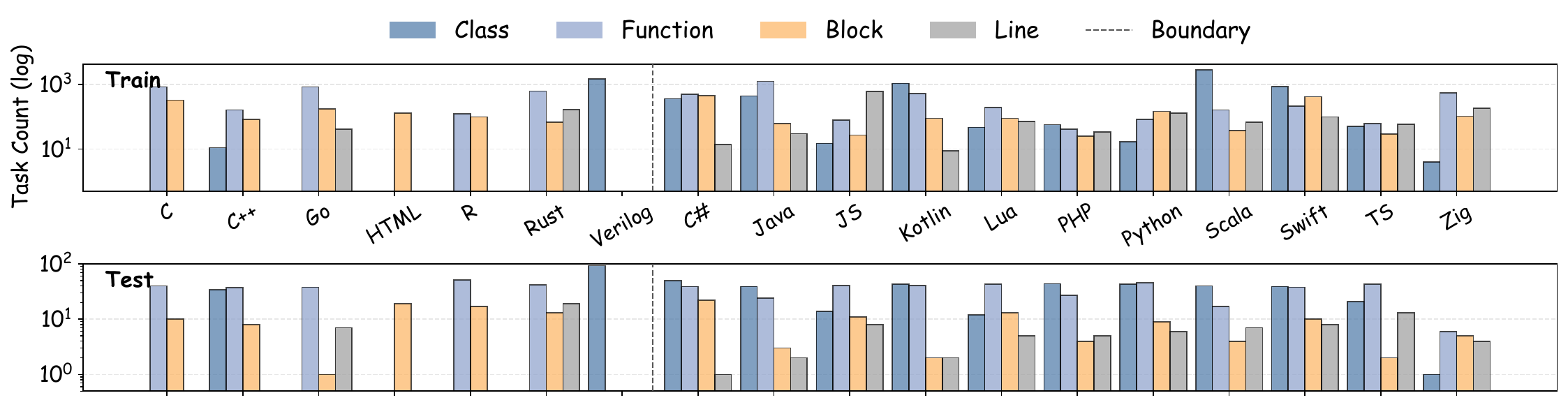}
        \caption{Task count of $\mathcal{D}_{\text{t}}$ and $\mathcal{D}_{\text{e}}$. The Y-axis is logarithmic; the left side of the dashed line is a partial-granularity group, and the right side is a full-granularity group. The same applies below.}
        \label{fig:taskcount}
\end{figure*}

\begin{figure*}[!tb]
\centering
        \includegraphics[width=1.0\textwidth]{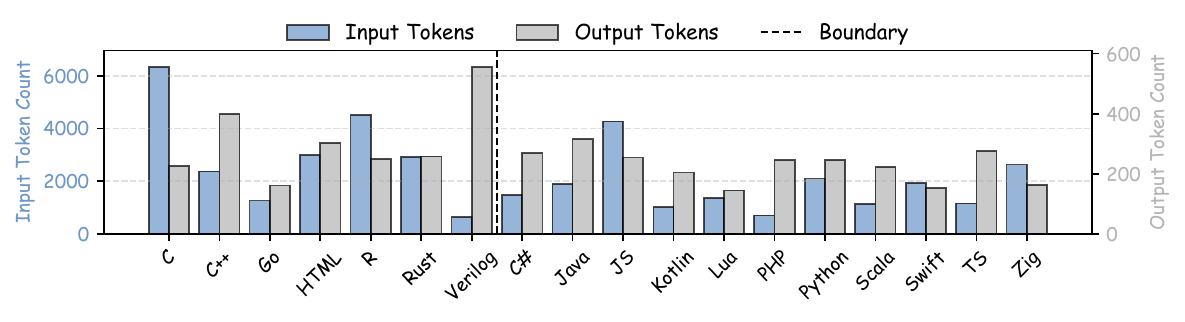}
        \caption{Task input and output statistics of $\mathcal{D}_{\text{t}}$ and $\mathcal{D}_{\text{e}}$.}
        \label{fig:tasktoken}
\end{figure*}

\begin{table*}[!htb]
\centering
\scalebox{0.8}{
\setlength{\tabcolsep}{5pt}
\begin{tabular}{llcccccc}
\toprule
\textbf{Frameworks} & \textbf{Primary Task} & \textbf{Class} & \textbf{Function} & \textbf{Block} & \textbf{Line} & \textbf{Cross-file} & \textbf{Language} \\
\midrule
HumanEval        & Generation              & -- & \checkmark & -- & --  & --          & Python  \\
MBPP             & Generation              & -- & \checkmark & -- & --  & --          & Python  \\
MultiPL\mbox{-}E & Generation (translated) & -- & \checkmark & -- & --  & --          & 18 \\
CrossCodeEval    & Repo-level Completion   & -- & -- & -- & --  & \checkmark      & 4  \\
M2RC\mbox{-}Eval & Repo-level Completion   & -- & -- & -- & --  & \checkmark      & 18 \\
CodeEditorBench  & Editing/Refinement      & -- & -- & -- & -- & --       & 3  \\
CanItEdit        & Instructional Editing   & -- & -- & -- & --  & --       & Python  \\
\rowcolor{gray!10}
\textbf{\benchmark{} (Ours)} & \textbf{Generation (multi-granularity)} &
\textbf{\checkmark} & \textbf{\checkmark} & \textbf{\checkmark} & \textbf{\checkmark}  &
\textbf{retrieval} & \textbf{18} \\
\bottomrule
\end{tabular}
}
\vspace{-2mm}
\caption{Comparison of code generation frameworks}
\vspace{-4mm}
\label{tab:benchmark-comparison}
\end{table*}

\paragraph{DataSet Construction and Quality Control.}
To reduce pretraining contamination, we build $R_{\text{e}}$ from GitHub repositories created or last updated after January 1, 2024. Because data volume varies widely across languages, we split the 18 languages into full-granularity and partial-granularity groups. Languages in the full-granularity group (e.g., Python, Java) have test cases at all four granularities, whereas languages in the partial-granularity group (e.g., Verilog, HTML) lack test cases at one or more granularities. Unlike the scale-oriented training dataset $\mathcal{D}_{\text{t}}$, the evaluation dataset $\mathcal{D}_{\text{e}}$ follows a quality-first pipeline. We first use the strong reasoning model DeepSeek-R1~\citep{deepseek_r1}, denoted $G_{\text{r}}$, to generate candidate tasks and apply the same $S$-based filter to obtain a provisional set $\mathcal{D}'_{\text{e}}$. Then, a team of 10 graduate and doctoral students with solid programming backgrounds reviews, tests, and refines each candidate, ensuring semantic correctness, complete context, and appropriate difficulty. The final $\mathcal{D}_{\text{e}}$ contains 1,286 carefully validated test instances, and constructing this set takes about 28–36 hours per language, compared with 6–8 hours per language for the automated training pipeline.

\paragraph{Comparison.} \autoref{tab:benchmark-comparison} compares \benchmark{} with mainstream code generation frameworks, highlighting the value of its multi-granularity design. Existing frameworks show critical limitations: HumanEval/MBPP support only single-granularity (Function-level) generation with 1 language~\citep{humaneval,mbpp}; CrossCodeEval/M2RC-Eval enables cross-file completion but lacks granularity distinction~\citep{crosscodeeval,m2rceval}; CodeEditorBench/CanItEdit focuses on editing but omits cross-file/multilingual support~\citep{codeeditorbench,editevaluatingabilitylarge}; and MultiPL-E still restricts to single-granularity~\citep{multipl_e}. These gaps directly motivate our \benchmark{} design, along with the associated \instruction{}, which jointly address the lack of multi-granularity, cross-file, and multilingual support.

\begin{table*}[!tb]
  \centering
  \resizebox{1.0\textwidth}{!}{%
  \begin{tabular}{lr|cc|ccc|ccc|cc|cc|ccc|c|cccc}
  \hline
  \multirow{2}{*}{Model} & \multirow{2}{*}{Size} &
  \multicolumn{2}{c|}{C} & 
  \multicolumn{3}{c|}{C++} & 
  \multicolumn{3}{c|}{Go} & 
  \multicolumn{2}{c|}{Html} & 
  \multicolumn{2}{c|}{R} & 
  \multicolumn{3}{c|}{Rust} & 
  \multicolumn{1}{c|}{\small Verilog} & 
  \multicolumn{4}{c}{Average}
  \\
  & & 
  \textbf{F} & \textbf{B}  & \textbf{C} & \textbf{F} & \textbf{B} & \textbf{F} & \textbf{B} & \textbf{L} & \textbf{B} & \textbf{L} & \textbf{F} & \textbf{B} & \textbf{F} & \textbf{B} & \textbf{L} &  \textbf{C} & \textbf{C} & \textbf{F} & \textbf{B}  & \textbf{L}  \\
  \noalign{\global\aboverulesep=0pt \global\belowrulesep=0pt}
  \midrule
  \rowcolor{ClosedColor}\multicolumn{22}{c}{\textit{\textbf{Closed-Source LLMs}}}\\[-0.1ex]
  \midrule
  \small Claude-3-7-Sonnet & \faLock{}
  & 26.2 & 31.9
  & 7.5 & \textbf{26.0} & 41.4
  & \textbf{26.6} & 19.0 & 20.8
  & 42.0 & 35.6
  & 21.2 & 19.8
  & 28.7 & \textbf{35.0} & \small \textbf{59.0}
  & \textbf{13.0}
  & 7.5 & 25.7 & 31.5 & 38.5
  \\
  \small Claude-4-Sonnet & \small \faLock{}
  & 28.6 & 31.6
  & \textbf{14.1} & 24.5 & 38.1
  & 24.4 & 18.0 & 21.9
  & 35.9 & 38.4
  & 20.6 & 22.9
  & 27.8 & 27.6 & 29.4
  & 10.4
  & \textbf{14.1} & 25.2 & 29.0 & 29.9
  
  \\
  \small o4-mini & \small \faLock{}
   & 27.9 & 30.9
  & 1.0 & 23.6 & 31.0
   & 26.0 & 14.0 & 44.7
   & \textbf{49.1} & \textbf{51.5}
   & 17.9 & 19.9
   & 31.3 & 24.9 & 29.2
   & 4.3
   & 11.6 & 22.6 & 26.8 & 37.6
  \\
  \small gpt-4o-2024-11-20 & \small \faLock{}
   & 3.4 & 5.5
   & 1.6 & 10.9 & 3.4
   & 8.4 & 12.0 & 12.3
   & 12.2 & 7.5
   & 20.5 & 10.0
   & 12.7 & 19.9 & 19.9
   & 0.2
   & 1.6 & 11.2 & 10.5 & 13.2
  \\
  \small o3-mini & \small \faLock{}
   & 15.5 & 4.3
   & 0.4 & 3.5 & 1.0
   & 2.5 & 8.0 & 5.2
  & 0.0 & 0.0
   & 13.9 & 10.4
   & 13.0 & 21.0 & 15.9
   & 0.0
    & 0.2 & 13.9 & 8.0 & 25.7
  \\
  \small gemini-2.5-pro & \small \faLock{}
   & \textbf{35.4} & \textbf{47.0}
   & 5.0 & 21.9 & \textbf{51.2}
   & 25.4 & \textbf{21.0} & \textbf{47.8}
   & 38.7 & 42.5
   & \textbf{25.9} & \textbf{29.1}
   & 29.2 & 32.4 & 36.5
  & 10.4
  & 5.0 & \textbf{27.6} & \textbf{36.6} & \textbf{42.3}
  \\
  \small gemini-2.5-flash & \small \faLock{}
   & 23.8 & 29.4
   & 10.8 & 24.9 & 38.0
   & 23.4 & 18.0 & 38.0
   & 35.3 & 43.5
   & 23.9 & 21.4
   & \textbf{33.3} & 31.5 & 28.9
   & 3.3
   & 10.8 & 25.9 & 28.9 & 36.8
  \\
  \midrule
   \rowcolor{OpenColor}\multicolumn{22}{c}{\textit{\textbf{Open-Source LLMs}}} \\
   \midrule
  \small Qwen3-0.6B-Chat & \small 0.6B
   & 7.0 & 3.2
   & 4.1 & 12.1 & 5.5
   & 8.0 & 6.0 & 3.9
   & 2.1 & 1.7
   & 12.5 & 6.5
   & 7.3 & 5.7 & 4.5
   & 1.5
   & 4.1 & 7.0 & 4.5 & 4.0
  \\
  \small Qwen3-0.6B-Think & \small 0.6B
   & 7.7 & 5.3
   & 4.0 & 12.1 & 2.7
   & 10.0 & - & 6.6
   & 3.2 & 2.8
   & 12.0 & 7.1
   & 6.0 & 6.8 & 6.8
   & 1.3
   & 3.5 & 7.9 & 5.1 & 5.5
  \\
  \small Qwen3-1.7B-Chat & \small 1.7B
   & 7.2 & 10.9
   & 3.3 & 4.5 & 5.4
   & 8.7 & 5.0 & 6.3
   & 11.9 & 9.5
   & 21.9 & 5.5
   & 9.7 & 9.4 & 11.4
   & 4.3
   & 3.3 & 8.7 & 7.6 & 6.3
  \\
  \small Qwen3-1.7B-Think & \small 1.7B
   & 6.6 & 7.2
   & 2.4 & 3.4 & 10.6
   & 7.8 & 4.0 & 11.8
   & 12.4 & 4.0
   & 22.6 & 6.6
   & 9.2 & 7.8 & 11.7
   & 4.7
   & 2.4 & 8.2 & 7.1 & 6.3
  \\
  \small DeepSeek-R1-Distill-Qwen-7B & \small 7B
   & 11.2 & 6.0
   & 2.1 & 10.3 & 6.8
   & 9.1 & 1.9 & 5.7
   & 15.2 & 7.2
   & 8.6 & 3.2
   & 12.4 & 4.6 & 6.5
   & 4.0
   & 2.1 & 10.3 & 6.3 & 6.5
  \\
  \small DeepSeek-R1-Distill-Qwen-14B & \small 14B
   & 8.9 & 10.7
   & 3.5 & 11.4 & 12.7
   & 11.0 & 16.0 & 11.6
   & 21.8 & 16.5
   & 18.8 & 10.4
   & 22.5 & 12.9 & 26.8
   & 3.5
   & 3.5 & 14.5 & 14.1 & 18.3
  \\
  \small Qwen3-14B-Chat & \small 14B
   & 14.5 & 17.2
  & 4.1 & 13.8 & 9.5
   & 17.5 & 8.0 & 20.1
   & 29.6 & 24.7
   & 21.6 & 16.7
   & 19.0 & 23.3 & 22.9
   & 6.7
   & 4.1 & 17.3 & 17.4 & 22.6
  \\
  \small Qwen3-14B-Think & \small 14B
   & 14.9 & 15.5
   & 5.6 & 15.6 & 20.5
   & 15.3 & 13.0 & 27.8
   & 33.7 & 24.1
   & 22.3 & 15.6
   & 21.4 & 22.2 & 22.0
   & 7.2
   & 5.6 & 17.9 & 20.1 & 24.6
  \\
  \small Qwen3-30B-A3B-Instruct & \small 3/30B
   & 8.2 & 28.2
   & 2.1 & 13.0 & 28.1
   & 12.6 & 9.0 & 19.3
   & 8.4 & 4.3
   & 21.5 & 10.4
   & 22.2 & 23.3 & 28.4
   & 0.8
   & 2.1 & 15.5 & 17.9 & 17.3
  \\
  \small Qwen3-30B-A3B-Think & \small 3/30B
   & \textbf{30.4} & 17.2
   & 7.1 & 17.4 & 28.0
   & 20.4 & 7.0 & 17.2
   & 6.9 & 11.1
   & 17.8 & 15.4
   & 23.6 & 24.6 & 28.4
   & 16.9
   & 7.1 & 21.9 & 16.5 & 18.9
  \\
  \small Qwen3-32B-Chat & \small 32B
   & 25.8 & 30.1
   & 6.2 & 23.3 & \textbf{31.5}
   & 20.4 & 10.0 & 33.5
   & 39.3 & 28.3
   & 20.7 & 17.8
   & 30.2 & 25.9 & 23.2
   & 7.6
   & 6.2 & 24.1 & 25.8 & 28.3
  \\
  \small Qwen-32B-Think & \small 32B
   & 24.4 & 29.6
  & 6.4 & 21.3 & 28.9
   & 19.9 & 9.0 & 23.1
   & 27.7 & 32.1
   & 20.6 & 16.8
   & 29.4 & 34.0 & 31.2
   & 7.5
   & 6.4 & 23.1 & 24.3 & 28.8
  \\
  \small DeepSeek-R1-Distill-Qwen-32B & \small 32B
   & 18.8 & 18.1
   & 4.1 & 15.6 & 25.5
   & 18.1 & 13.0 & 14.7
   & 36.2 & 28.9
   & 15.4 & 11.4
   & 21.5 & 14.8 & 33.5
   & 3.8
   & 4.1 & 17.9 & 19.8 & 25.7
  \\
  \small QwQ-32B & \small 32B
   & 22.5 & 15.3
   & 7.6 & 22.8 & 22.8
   & 21.0 & 9.0 & 22.3
   & 30.7 & 24.1
   & 19.6 & 16.9
   & 23.5 & 22.6 & 28.6
   & 7.4
   & 7.6 & 21.9 & 19.6 & 25.0
  \\
  \small DeepSeek-R1-Distill-Llama-70B & \small 70B
   & 11.5 & 24.7
   & 4.6 & 8.4 & 17.8
   & 14.3 & 6.0 & 13.9
   & 21.6 & 19.8
   & 19.2 & 13.0
   & 22.4 & 19.9 & 28.2
   & 2.2
   & 4.6 & 15.2 & 17.2 & 20.6
  \\
  \small Qwen3-235B-A22B-Think & \small 22/235B
   & 25.9 & 35.1
   & 6.5 & 21.3 & 25.0
   & 24.7 & 9.0 & 25.9
   & \textbf{40.7} & 33.5
   & 18.8 & 22.0
   & 20.8 & 26.5 & 27.6
   & 7.6
   & 6.5 & 22.3 & 26.4 & 29.0
  \\
  \small Qwen3-Coder-480B-A35B-Instruct & \small 35/480B
   & 23.5 & 43.1
   & 8.2 & 22.3 & 28.3
   & 24.1 & 14.0 & \textbf{49.7}
   & 38.8 & \textbf{46.8}
   & \textbf{24.9} & 20.4
   & 28.5 & \textbf{38.2} & 35.0
   & 4.7
   & 8.2 & 24.7 & \textbf{30.5} & \textbf{43.8}
  \\
  \small DeepSeek-R1 & \small 37/671B
   & 29.8 & 25.6
   & \textbf{30.1} & \textbf{28.6} & 26.2
   & \textbf{27.0} & 19.0 & 27.8
   & 33.8 & 26.9
   & 21.7 & \textbf{22.2}
   & \textbf{32.3} & 26.2 & 30.6
   & \textbf{24.1}
   & \textbf{30.1} & \textbf{27.9} & 25.5 & 28.4
  \\
  \small DeepSeek-V3 & \small 37/671B
   & 22.3 & \textbf{44.9}
   & 6.4 & 21.0 & 31.2
    & 23.4 & \textbf{27.0} & 39.4
    & 27.2 & 18.2
    & 24.2 & 18.0
   & 25.9 & 29.1 & \textbf{37.8}
   & 8.6
   & 6.4 & 23.4 & 29.6 & 31.8
  \\
  \small Qwen3-8B-Chat & \small 8B
  & 23.9 & 16.3
   & 8.3 & 18.1 & 22.0
   & 17.9 & 8.0 & 34.0
   & 18.6 & 13.6
   & 18.0 & 17.1
   & 24.9 & 19.2 & 31.9
   & 6.0
   & 8.3 & 20.6 & 16.9 & 26.5
  \\
  \small Qwen3-8B-Think & \small 8B
   & 22.6 & 16.2
   & 7.5 & 16.9 & 28.4
   & 18.5 & 8.0 & 14.8
   & 19.6 & 14.6
   & 20.6 & 19.8
   & 27.0 & 19.7 & 25.8
   & 8.0
   & 7.5 & 21.1 & 18.6 & 18.4
  \\
  \midrule
   \rowcolor{ProprietaryColor}\multicolumn{22}{c}{\textit{\textbf{Our Method}}} \\
  \midrule

  \small \codersft{} & \small 8B
   & 21.4 & 27.0
   & 7.6 & 18.1 & 21.8
   & 18.2 & 13.4 & 24.0
   & 32.9 & 36.4
   & \textbf{21.6} & 21.5
   & \textbf{30.0} & 20.5 & \textbf{35.5}
   & 7.7
   & 7.6 & 21.9 & 22.8 & 32.0
  \\ 
  \small \coderl{} & \small 8B
  & \textbf{24.6} & \textbf{27.8}
  & \textbf{7.8} & \textbf{21.9} & \textbf{23.9}
  & \textbf{19.8} & \textbf{15.0} & \textbf{36.7}
  & \textbf{35.5} & \textbf{40.7}
  & 21.2 & \textbf{23.5}
  & 25.2 & \textbf{23.8} & 31.7
  & \textbf{8.2}
  & \textbf{7.8} & \textbf{22.5} & \textbf{24.9} & \textbf{36.4}
  \\
  \bottomrule
  \end{tabular}
  } 
  \caption{Results on 7 partial-granularity languages.}
  \label{tab:7partiallangs}
  \vspace{-15pt}
  \end{table*}

\begin{table*}[!tb]
    \centering
    
    
    \resizebox{\textwidth}{!}{
    \begin{tabular}{lr|cccc|cccc|cccc|cccc|cccc|cccc}
    \toprule
    \multirow{2}{*}{Model} & \multirow{2}{*}{Size} 
    & \multicolumn{4}{c|}{C\#} 
    & \multicolumn{4}{c|}{Java} 
    & \multicolumn{4}{c|}{JS} 
    & \multicolumn{4}{c|}{Kotlin} 
    & \multicolumn{4}{c|}{Lua} 
    & \multicolumn{4}{c}{PHP} \\
    & & \textbf{C} & \textbf{F} & \textbf{B} & \textbf{L} & \textbf{C} & \textbf{F} & \textbf{B} & \textbf{L} & \textbf{C} & \textbf{F} & \textbf{B} & \textbf{L} & \textbf{C} & \textbf{F} & \textbf{B} & \textbf{L} &\textbf{C} & \textbf{F} & \textbf{B} & \textbf{L} &\textbf{C} & \textbf{F} & \textbf{B} & \textbf{L} \\
    \midrule
    \rowcolor{ClosedColor}\multicolumn{26}{c}{\textit{Closed-Source LLMs}}\\
    \midrule
     
    \small Claude-3-7-Sonnet & \small \faLock{}
    & \textbf{13.9} & 28.3 & 28.2 & 45.2
    & 20.4 & 31.4 & 10.5 & 24.4
    & 9.9 & 19.7 & 18.3 & 18.7
    & 21.3 & 25.2 & 11.6 & \textbf{42.9}
    & 30.9 & 19.6 & 24.5 & 11.7
    & \textbf{25.0} & 32.8 & 31.5 & 22.0
    \\
    \small Claude-4-Sonnet & \small \faLock{}
    & 12.6 & 29.0 & 33.1 & 45.2
    & 20.4 & 33.2 & 7.7 & 14.8
    & 10.1 & 20.8 & 18.7 & 18.7
    & 22.5 & 23.7 & 9.0 & 36.6
    & 30.2 & 21.9 & 28.0 & \textbf{18.5}
    & 20.2 & 30.8 & 26.8 & 16.8
    \\
    \small o4-mini & \small \faLock{}
    & 11.2 & 26.3 & 18.3 & 59.5
    & 18.0 & 26.1 & \textbf{39.5} & 25.5
    & 8.2 & 22.9 & 14.6 & 17.8
    & \textbf{30.0} & 31.1 & 20.5 & 40.8
    & 24.3 & 21.2 & 31.1 & 4.1
    & 12.6 & 24.9 & 38.8 & \textbf{26.0}
    \\
    \small gpt-4o-2024-11-20 & \small \faLock{}
    & 0.0 & 0.0 & 1.5 & 66.7
    & 3.7 & 8.6 & 7.1 & 18.4
    & 4.7 & 16.4 & 3.3 & 11.7
    & 6.4 & 5.6 & 32.2 & 0.0
    & 17.2 & 4.9 & 13.8 & 7.2
    & 0.4 & 10.4 & 1.5 & 19.1
    \\
    \small o3-mini & \small \faLock{}
    & 0.0 & 13.4 & 9.6 & \textbf{81.0}
    & 0.0 & 13.4 & 8.5 & 15.9
    & 5.8 & 14.3 & 7.5 & 2.0
    & 1.1 & 0.5 & \textbf{73.4} & 40.8
    & 18.3 & 11.8 & 18.7 & 0.0
    & 0.0 & 15.9 & 35.8 & 9.5
    \\
    \small gemini-2.5-pro & \small \faLock{}
    & 12.7 & \textbf{31.7} & \textbf{36.8} & 67.3
    & \textbf{24.1} & \textbf{35.0} & 19.1 & \textbf{27.8}
    & 10.4 & \textbf{26.4} & \textbf{23.9} & 39.9
    & 23.7 & 28.9 & 45.2 & 33.0
    & \textbf{40.9} & \textbf{23.0} & \textbf{36.6} & 13.8
    & 21.3 & \textbf{37.2} & \textbf{46.0} & 21.6
    \\
    \small gemini-2.5-flash & \small \faLock{}
    & 9.2 & 29.5 & 27.6 & 0.0
    & 13.5 & 28.3 & 11.4 & 26.6
    & \textbf{10.9} & 23.4 & 20.8 & \textbf{41.6}
    & 21.7 & \textbf{31.3} & 29.1 & 28.7
    & 28.9 & 22.3 & 28.9 & 11.7
    & 10.7 & 22.8 & 21.6 & 11.9
    \\
    \midrule
    \rowcolor{OpenColor}\multicolumn{26}{c}{\textit{Open-Source LLMs}} \\
    \midrule
    \small Qwen3-0.6B-Chat & \small 0.6B
    & 1.1 & 7.1 & 1.6 & 2.4
    & 2.7 & 6.0 & 3.2 & 2.1
    & 1.3 & 3.9 & 5.2 & 16.7
    & 5.6 & 8.0 & 6.1 & 5.8
    & 8.8 & 7.9 & 8.3 & 5.1
    & 1.7 & 5.2 & 4.5 & 5.4
    \\
    \small Qwen3-0.6B-Think & \small 0.6B
    & 1.2 & 6.7 & 3.7 & 2.4
    & 2.7 & 4.9 & 4.1 & 3.2
    & 1.3 & 6.4 & 3.8 & 9.8
    & 6.1 & 6.3 & 6.4 & 0.0
    & 8.2 & 7.4 & 8.7 & 3.8
    & 1.7 & 4.6 & 2.6 & 9.8
    \\
    \small Qwen3-1.7B-Chat & \small 1.7B
    & 3.4 & 9.0 & 8.2 & 0.0
    & 4.4 & 7.2 & 8.6 & 3.6
    & 7.5 & 9.9 & 6.0 & 2.3
    & 10.0 & 6.4 & 1.7 & 0.0
    & 8.9 & 6.4 & 11.0 & 7.5
    & 5.6 & 7.3 & 9.7 & 5.2
    \\
    \small Qwen3-1.7B-Think & \small 1.7B
    & 6.2 & 7.4 & 13.4 & 9.4
    & 5.0 & 8.5 & 14.0 & 3.4
    & 6.3 & 7.2 & 4.6 & 4.7
    & 10.9 & 8.0 & 2.1 & 1.8
    & 12.9 & 8.2 & 8.5 & 5.7
    & 8.5 & 9.2 & 13.1 & 5.7
    \\
    \small DeepSeek-R1-Distill-Qwen-7B & \small 7B
    & 5.2 & 9.8 & 10.1 & 0.0
    & 6.9 & 9.5 & 7.5 & 3.6
    & 6.0 & 9.3 & 4.6 & 4.3
    & 6.5 & 6.1 & 9.1 & 0.0
    & 9.6 & 7.9 & 7.3 & 7.9
    & 7.6 & 9.1 & 20.8 & 9.4
    \\
    \small DeepSeek-R1-Distill-Qwen-14B & \small 14B
    & 8.0 & 15.9 & 16.9 & 5.8
    & 9.5 & 17.5 & 15.6 & 9.3
    & 4.5 & 13.7 & 16.8 & 16.0
    & 11.5 & 16.4 & 12.5 & \textbf{49.0}
    & 21.4 & 13.2 & 18.1 & 13.2
    & 7.3 & 15.1 & 21.4 & 25.9
    \\
    \small Qwen3-14B-Chat & \small 14B
    & 9.1 & 23.3 & 26.3 & 9.4
    & 16.1 & 24.8 & 13.4 & 18.5
    & 9.1 & 18.6 & 16.6 & 25.7
    & 16.9 & 18.7 & 9.0 & 34.7
    & 17.0 & 17.8 & 17.1 & 13.2
    & 15.4 & 20.5 & 28.1 & 14.9
    \\
    \small Qwen3-14B-Think & \small 14B
    & 11.9 & 23.1 & 22.1 & 9.4
    & 16.0 & 25.8 & 18.6 & 14.4
    & 9.2 & 16.2 & 18.4 & 15.6
    & 16.1 & 18.8 & 6.5 & 34.7
    & 19.5 & 17.6 & 19.0 & 10.3
    & 14.5 & 21.0 & 19.7 & 16.9
    \\
    \small Qwen3-30B-A3B & \small 3/30B
    & 4.6 & 21.2 & 28.3 & 61.9
    & 8.2 & 26.2 & 21.1 & 19.4
    & 6.9 & 14.6 & 18.1 & 17.4
    & 16.0 & 14.7 & 36.9 & 22.4
    & 23.2 & 12.5 & 20.2 & 11.8
    & 2.2 & 18.9 & 24.5 & 16.0
    \\
    \small Qwen3-30B-A3B-Think & \small 3/30B
    & 22.7 & 27.9 & 19.9 & 50.0
    & 14.3 & 28.5 & 12.6 & 15.2
    & 6.8 & 23.3 & 21.1 & 22.9
    & 22.4 & 20.0 & 26.3 & 42.9
    & 19.6 & 13.8 & 13.1 & 7.7
    & 20.0 & 21.5 & 35.0 & 13.4
    \\
    \small Qwen3-32B-Chat & \small 32B
    & 8.2 & 24.4 & 21.7 & 27.1
    & 14.6 & 26.1 & 13.7 & 15.4
    & 8.3 & 21.6 & 18.0 & 19.2
    & 18.7 & 20.5 & 10.0 & 30.3
    & 32.5 & 19.7 & 15.2 & 9.5
    & 16.5 & 20.3 & 27.8 & 30.1
    \\
    \small Qwen-32B-Think & \small 32B
    & 10.5 & 23.9 & 25.4 & 23.1
    & 15.8 & 24.2 & 18.2 & 18.2
    & 8.5 & 19.7 & 18.5 & 22.4
    & 17.6 & 23.4 & 3.5 & 28.8
    & 25.7 & 19.3 & 16.6 & 8.0
    & 15.5 & 21.2 & 27.8 & 31.3
    \\
    \small DeepSeek-R1-Distill-Qwen-32B & \small 32B
    & 5.1 & 19.2 & 18.6 & 6.5
    & 9.7 & 15.9 & 1.4 & 12.9
    & 5.6 & 16.4 & 18.7 & 9.7
    & 10.8 & 15.6 & 34.4 & 27.1
    & 28.4 & 16.4 & 16.4 & 13.1
    & 8.0 & 19.1 & 29.3 & 18.0
    \\
    \small QwQ-32B & \small 32B
    & 9.2 & 19.3 & 17.8 & 38.2
    & 14.2 & 23.3 & 11.7 & 15.8
    & 8.2 & 19.7 & 15.2 & 27.0
    & 13.7 & 18.5 & 16.2 & 20.4
    & 22.3 & 18.4 & 20.3 & 9.7
    & 18.5 & 17.4 & 16.1 & 17.0
    \\
    \small DeepSeek-R1-Distill-Llama-70B & \small 70B
    & 6.5 & 13.1 & 19.9 & 52.4
    & 12.9 & 17.6 & 17.7 & 13.8
    & 6.1 & 18.6 & 12.7 & 9.7
    & 10.4 & 13.9 & 36.4 & 12.6
    & 30.8 & 15.0 & 15.8 & 11.1
    & 7.5 & 14.9 & 33.6 & 10.1
    \\
    \small Qwen3-235B-A22B-Think & \small 22/235B
    & 8.9 & 29.4 & 23.6 & 59.5
    & 16.9 & 26.1 & 23.0 & 24.9
    & 7.6 & \textbf{25.9} & 17.4 & 24.0
    & 17.9 & 22.1 & \textbf{46.9} & 22.0
    & 29.3 & 21.2 & 22.3 & 15.0
    & 13.3 & 20.3 & 35.1 & 15.8
    \\
    \small Qwen3-Coder-480B-A35B-Instruc & \small 35/480B
    & 12.2 & \textbf{35.0} & \textbf{31.4} & \textbf{71.4}
    & 21.7 & 30.0 & 6.1 & \textbf{26.9}
    & 11.7 & 23.9 & 19.7 & \textbf{36.6}
    & 20.1 & 16.1 & 10.5 & 36.9
    & \textbf{37.2} & 21.8 & 26.4 & 14.2
    & 17.9 & \textbf{28.6} & \textbf{54.3} & \textbf{31.8}
    \\
    \small DeepSeek-R1 & \small 37/671B
    & \textbf{31.0} & 30.0 & 29.3 & 47.6
    & \textbf{30.0} & \textbf{33.2} & 15.8 & 24.9
    & \textbf{19.9} & 24.3 & \textbf{21.5} & 32.0
    & \textbf{26.5} & \textbf{32.0} & 27.1 & 32.0
    & 31.0 & 21.8 & \textbf{30.7} & 15.4
    & \textbf{35.2} & 27.6 & 34.8 & 19.6
    \\
    \small DeepSeek-V3 & \small 37/671B
    & 9.9 & 22.2 & 25.8 & 42.4
    & 16.1 & 22.4 & \textbf{24.0} & 13.7
    & 9.5 & 20.9 & 16.7 & 31.9
    & 20.3 & 17.2 & 6.2 & 28.2
    & 35.1 & \textbf{22.3} & 28.5 & \textbf{21.6}
    & 18.7 & 23.2 & 30.3 & 17.3
    \\
    \small Qwen3-8B Chat & \small 8B
    & 8.0 & 22.0 & 21.3 & 9.4
    & 10.6 & 22.0 & 20.7 & 15.2
    & 8.3 & 18.2 & 15.0 & 22.1
    & 16.1 & 20.9 & 20.2 & 42.9
    & 21.3 & 15.8 & 11.9 & 11.6
    & 14.7 & 17.3 & 36.2 & 10.2
    \\
    \small Qwen3-8B Think & \small 8B
    & 12.3 & 23.6 & 21.0 & 9.4
    & 11.8 & 22.2 & 22.3 & 13.8
    & 7.6 & 17.5 & 12.9 & 19.9
    & 16.6 & 18.7 & 30.8 & 28.7
    & 23.0 & 17.5 & 14.3 & 11.9
    & 15.1 & 16.8 & 18.2 & 13.7
    \\
    \midrule
    \rowcolor{ProprietaryColor}\multicolumn{26}{c}{\textit{Our Method}}\\
    \midrule
    \small \codersft{} & \small 8B
    & \textbf{13.5} & \textbf{24.6} & 23.9 & 32.4
    & 18.1 & 27.4 & 14.2 & 13.9
    & \textbf{10.0} & \textbf{19.3} & \textbf{19.7} & 20.5
    & 16.8 & \textbf{19.8} & 35.4 & 32.5
    & 27.7 & 19.4 & \textbf{27.6} & 13.6
    & \textbf{17.5} & 23.8 & 31.7 & 36.2
    \\
    \small \coderl{} & \small 7B
    & \textbf{13.5} & 21.7 & \textbf{27.9} & \textbf{73.8}
    & \textbf{19.0} & \textbf{31.1} & \textbf{21.1} & \textbf{17.7}
    & 9.9 & 18.1 & 19.0 & \textbf{27.3}
    & \textbf{17.6} & 19.3 & \textbf{36.6} & \textbf{33.6}
    & \textbf{31.8} & \textbf{19.6} & \textbf{27.6} & \textbf{16.2}
    & 17.0 & \textbf{24.8} & \textbf{34.5} & \textbf{36.5}
    \\
    \bottomrule

    
    \multirow{2}{*}{Model} & \multirow{2}{*}{Size} 
    & \multicolumn{4}{c|}{\textbf{Python}}
    & \multicolumn{4}{c|}{\textbf{Scala}}
    & \multicolumn{4}{c|}{\textbf{Swift}}
    & \multicolumn{4}{c|}{\textbf{TS}}
    & \multicolumn{4}{c|}{\textbf{Zig}}
    & \multicolumn{4}{c}{\textbf{Average}} \\
    && \textbf{C} & \textbf{F} & \textbf{B} & \textbf{L} & \textbf{C} & \textbf{F} & \textbf{B} & \textbf{L} & \textbf{C} & \textbf{F} & \textbf{B} & \textbf{L} & \textbf{C} & \textbf{F} & \textbf{B} & \textbf{L} &\textbf{C} & \textbf{F} & \textbf{B} & \textbf{L} &\textbf{C} & \textbf{F} & \textbf{B} & \textbf{L} \\
    \midrule
    \rowcolor{ClosedColor}\multicolumn{26}{c}{\textit{Closed-Source LLMs}}\\
    \midrule
    \small Claude-3-7-Sonnet & \small \faLock{}
    & 16.8 & 25.0 & 25.2 & 28.8
    & 16.3 & 25.7 & 11.2 & 19.1
    & 20.5 & 24.5 & 18.9 & 17.4
    & 17.2 & 24.4 & 30.3 & 31.2
    & 18.8 & \textbf{35.2} & 23.4 & 38.7
    & 19.2 & 25.7 & 21.0 & 26.1
    \\
    \small Claude-4-Sonnet & \small \faLock{}
    & 18.2 & 22.1 & 22.5 & 34.5
    & 15.9 & \textbf{31.0} & 14.2 & 24.3
    & \textbf{26.5} & 28.4 & 23.0 & \textbf{33.6}
    & 16.4 & 21.6 & 34.6 & 29.7
    & \textbf{20.7} & 19.7 & 21.3 & 29.9
    & 19.3 & 26.3 & 21.8 & 27.3
    \\
    \small o4-mini & \small \faLock{}
    & 5.0 & 17.6 & 26.5 & \textbf{50.2}
    & \textbf{32.7} & 28.6 & 10.6 & 20.8
    & 24.2 & 17.4 & 19.4 & 18.6
    & 19.2 & 19.0 & 25.6 & 28.1
    & 0.1 & 29.3 & 15.9 & 8.9
    & 18.7 & 23.0 & 24.3 & 31.1
    \\
    \small gpt-4o-2024-11-20 & \small \faLock{}
    & 1.7 & 7.2 & 1.1 & 12.7
    & 6.0 & 10.1 & 7.6 & 8.9
    & 0.1 & 4.6 & 14.7 & 11.5
    & 10.9 & 6.8 & 0.0 & 8.3
    & 11.1 & 8.1 & 10.6 & 10.9
    & 5.1 & 7.5 & 8.3 & 16.5
    \\
    \small o3-mini & \small \faLock{}
    & 2.3 & 4.0 & 11.1 & 8.0
    & 4.0 & 0.0 & 8.3 & 28.6
    & 0.1 & 2.1 & 20.1 & 0.0
    & 3.0 & 4.9 & 0.0 & 16.6
    & 12.3 & 0.0 & 13.6 & 26.4
    & 3.4 & 8.5 & 9.4 & 14.5
    \\
    \small gemini-2.5-pro & \small \faLock{}
    & \textbf{20.3} & \textbf{29.8} & \textbf{27.3} & 25.9
    & 19.5 & 27.9 & \textbf{30.1} & \textbf{50.8}
    & 22.5 & 28.3 & \textbf{32.2} & 21.8
    & \textbf{24.4} & \textbf{28.6} & 37.4 & \textbf{37.4}
    & 1.5 & 24.7 & \textbf{48.3} & \textbf{46.2}
     & \textbf{22.0} & \textbf{29.7} & \textbf{33.5} & \textbf{33.9}
    \\
    \small gemini-2.5-flash & \small \faLock{}
    & 15.3 & 23.0 & 26.3 & 23.9
    & 17.0 & 26.6 & 20.7 & 26.2
    & 18.9 & \textbf{28.5} & 23.4 & 25.6
    & 14.5 & 24.9 & \textbf{38.6} & 32.9
    & 19.9 & 32.3 & 25.3 & 37.6
    & 16.1 & 26.1 & 24.8 & 25.7
    \\
    \midrule
    \rowcolor{OpenColor}\multicolumn{26}{c}{\textit{Open-Source LLMs}}\\
    \midrule
    \small Qwen3-0.6B-Chat & \small 0.6B
    & 4.9 & 8.2 & 5.2 & 7.0
    & 6.6 & 6.8 & 5.0 & 5.3
    & 4.1 & 4.8 & 4.5 & 10.3
    & 3.7 & 6.0 & 0.4 & 5.8
    & 0.8 & 4.0 & 7.0 & 2.4
    & 3.8 & 6.1 & 4.8 & 6.1
    \\
    \small Qwen3-0.6B-Think & \small 0.6B
    & 5.0 & 8.5 & 4.4 & 7.7
    & 5.8 & 5.7 & 6.3 & 8.5
    & 3.6 & 4.3 & 4.6 & 6.7
    & 5.4 & 5.8 & 0.0 & 3.8
    & 0.2 & 2.2 & 6.3 & 0.9
    & 3.7 & 5.7 & 5.1 & 5.4
    \\
    \small Qwen3-1.7B-Chat & \small 1.7B
    & 11.4 & 10.4 & 5.8 & 4.3
    & 5.5 & 2.9 & 4.6 & 6.9
    & 5.8 & 6.7 & 7.2 & 3.6
    & 4.9 & 9.6 & 14.6 & 4.7
    & 1.4 & 4.9 & 16.4 & 7.8
    & 6.3 & 7.4 & 8.4 & 5.3
    \\
    \small Qwen3-1.7B-Think & \small 1.7B
    & 12.1 & 10.4 & 7.4 & 7.4
    & 5.1 & 5.1 & 6.3 & 6.2
    & 8.3 & 6.4 & 8.5 & 7.5
    & 5.9 & 10.6 & 16.4 & 6.3
    & 8.7 & 6.4 & 13.5 & 8.0
    & 7.9 & 7.7 & 9.3 & 6.9
    \\
    \small DeepSeek-R1-Distill-Qwen-7B & \small 7B
    & 8.4 & 10.7 & 7.2 & 6.3
    & 6.8 & 14.4 & 9.4 & 4.8
    & 8.9 & 7.7 & 7.5 & 4.6
    & 6.9 & 8.7 & 0.8 & 8.0
    & 0.0 & 2.1 & 11.6 & 0.0
    & 7.3 & 9.3 & 8.4 & 6.1
    \\
    \small DeepSeek-R1-Distill-Qwen-14B & \small 14B
    & 9.9 & 13.9 & 13.6 & 16.6
    & 9.8 & 15.5 & 10.0 & 9.6
    & 8.7 & 12.1 & 15.8 & 11.8
    & 7.1 & 13.0 & 16.0 & 18.3
    & 18.1 & 6.8 & \textbf{30.9} & 8.2
    & 9.8 & 14.6 & 15.7 & 17.6
    \\
    \small Qwen3-14B-Chat & \small 14B
    & 16.0 & 19.0 & 19.2 & \textbf{39.0}
    & 11.1 & 24.7 & 12.4 & 15.1
    & 15.3 & 14.1 & 18.3 & 4.7
    & 13.1 & 14.1 & 6.0 & 24.3
    & 18.2 & 9.1 & 16.0 & 14.8
    & 13.9 & 19.7 & 16.6 & 20.0
    \\
    \small Qwen3-14B-Think & \small 14B
    & 14.5 & 20.6 & 12.4 & 27.1
    & 11.9 & 21.5 & 10.5 & 19.3
    & 14.1 & 14.2 & 15.7 & 10.1
    & 10.5 & 14.1 & 7.9 & 27.7
    & 19.8 & 10.3 & 20.0 & 25.4
    & 13.8 & 19.3 & 15.1 & 18.6
    \\
    \small Qwen3-30B-A3B & \small 3/30B
    & 11.5 & 14.3 & 19.3 & 20.0
    & 8.1 & 13.0 & 11.5 & 15.6
    & 11.5 & 17.8 & 20.8 & 12.5
    & 11.5 & 16.8 & 27.0 & 23.8
    & \textbf{22.9} & 6.9 & 27.6 & 20.5
    & 10.4 & 17.0 & 22.8 & 22.1
    \\
    \small Qwen3-30B-A3B-Think & \small 3/30B
    & 10.8 & 22.0 & 23.5 & 38.2
    & 13.6 & 24.0 & 16.1 & 12.1
    & 16.5 & 14.6 & 17.9 & 21.5
    & 15.9 & 19.6 & 16.6 & 34.2
    & 1.4 & \textbf{31.7} & 25.0 & 14.0
    & 16.3 & 21.5 & 20.2 & 25.8
    \\
    \small Qwen3-32B-Chat & \small 32B
    & 17.9 & 23.0 & 17.7 & 30.5
    & 14.0 & 27.1 & 15.5 & 24.7
    & 16.4 & 25.4 & 18.1 & 19.9
    & 11.5 & 22.2 & 24.9 & 30.1
    & 13.4 & 24.6 & 21.7 & 21.2
    & 15.9 & 23.0 & 18.3 & 23.7
    \\
    \small Qwen3-32B-Think & \small 32B
    & 15.4 & 21.3 & 16.4 & 31.9
    & 15.3 & 23.1 & 11.5 & 21.7
    & 16.9 & \textbf{29.0} & 21.4 & 15.4
    & 11.4 & 19.2 & \textbf{42.3} & 26.8
    & 9.3 & 23.3 & 20.2 & 34.9
    & 15.3 & 22.4 & 20.1 & 22.8
    \\
    \small DeepSeek-R1-Distill-Qwen-32B & \small 32B
    & 12.3 & 18.5 & 14.2 & 15.0
    & 7.1 & 13.4 & 15.5 & 18.6
    & 9.5 & 12.3 & 20.7 & 18.8
    & 8.8 & 12.8 & 13.4 & 25.7
    & 9.0 & 13.6 & 20.3 & 20.5
    & 10.5 & 16.0 & 18.3 & 16.5
    \\
    \small QwQ-32B & \small 32B
    & 15.4 & 19.9 & 19.6 & 34.2
    & 10.9 & 28.3 & 16.3 & 25.6
    & 15.2 & 20.4 & 19.1 & 16.2
    & 10.6 & 19.7 & 33.9 & 25.8
    & 6.1 & 13.0 & 9.7 & 25.5
    & 13.8 & 20.5 & 18.6 & 23.0
    \\
    \small DeepSeek-R1-Distill-Llama-70B & \small 70B
    & 12.3 & 19.8 & 16.0 & 23.8
    & 7.6 & 13.4 & 21.7 & 21.4
    & 15.7 & 13.2 & 13.9 & 4.8
    & 8.4 & 12.6 & 13.4 & 17.8
    & 1.5 & 9.6 & 14.3 & 20.9
    & 11.8 & 15.2 & 20.1 & 17.8
    \\
    \small Qwen3-235B-A22B-Think & \small 22/235B
    & 16.4 & 22.3 & 19.9 & 35.9
    & 13.2 & 25.5 & 15.7 & 31.0
    & 17.4 & 26.4 & 19.8 & 22.1
    & 13.3 & 20.5 & 33.8 & 26.1
    & 1.5 & 23.8 & 22.4 & 30.0
    & 15.4 & 24.0 & \textbf{25.8} & 27.6
    \\
    \small Qwen3-Coder-480B-A35B-Instruct & \small 35/480B
    & 12.7 & 23.2 & \textbf{32.5} & 26.2
    & 13.5 & 27.5 & 19.4 & 26.7
    & 19.6 & 23.5 & 20.0 & 23.3
    & 16.9 & 22.4 & 31.8 & 34.1
    & 7.4 & 19.8 & 13.2 & \textbf{39.3}
    & 18.4 & 25.2 & 25.2 & \textbf{32.8}
    \\
    \small DeepSeek-R1 & \small 37/671B
    & \textbf{25.6} & \textbf{27.4} & 20.8 & 33.4
    & \textbf{22.3} & \textbf{33.4} & \textbf{24.6} & 27.0
    & \textbf{21.4} & 27.0 & \textbf{25.5} & \textbf{28.8}
    & \textbf{30.9} & \textbf{30.0} & 25.3 & 30.2
    & 11.0 & 24.2 & 26.8 & 29.0
    & \textbf{27.4} & \textbf{28.7} & 25.5 & 29.1
    \\
    \small DeepSeek-V3 & \small 37/671B
    & 13.7 & 20.4 & 22.1 & 36.4
    & 16.6 & 25.8 & 19.7 & \textbf{37.2}
    & 17.7 & 25.0 & 23.6 & 14.7
    & 19.3 & 20.4 & 28.4 & \textbf{37.0}
    & 16.1 & 11.9 & 22.9 & 37.1
    & 17.7 & 22.0 & 22.5 & 28.0
    \\
    \small Qwen3-8B Chat & \small 8B
    & 12.6 & 16.7 & 13.9 & 23.8
    & 13.4 & 21.3 & 13.5 & 18.8
    & 17.3 & 15.7 & 14.8 & 22.4
    & 12.6 & 20.1 & 16.4 & 28.9
    & 4.3 & 10.1 & 13.5 & 11.6
    & 13.5 & 19.0 & 18.4 & 20.5
    \\
    \small Qwen3-8B Think & \small 8B
    & 12.5 & 17.8 & 16.8 & 14.8
    & 14.0 & 22.9 & 14.2 & 11.1
    & 20.6 & 14.3 & 15.2 & 15.3
    & 14.1 & 18.2 & 22.9 & 22.0
    & 8.0 & 13.3 & 14.7 & 13.5
    & 14.8 & 19.0 & 18.9 & 16.1
    \\
    \midrule
    \rowcolor{ProprietaryColor}\multicolumn{26}{c}{\textit{Our Method}}\\
    \midrule
    \small \codersft{} & \small 8B
    & 12.8 & 20.9 & \textbf{21.2} & 38.2
    & \textbf{12.8} & 19.3 & 9.7 & 22.7
    & 17.4 & 22.6 & 13.8 & \textbf{28.7}
    & \textbf{14.4} & 20.8 & 21.0 & \textbf{27.9}
    & \textbf{16.3} & \textbf{21.8} & \textbf{32.9} & \textbf{18.9}
    & 16.1 & 21.8 & 21.8 & 26.7
    \\
    \small \coderl{} & \small 8B
    & \textbf{14.3} & \textbf{22.0} & 19.2 & \textbf{42.5}
    & 12.6 & \textbf{21.9} & \textbf{13.0} & \textbf{23.0}
    & \textbf{18.4} & \textbf{23.0} & \textbf{24.5} & 26.9
    & 14.3 & \textbf{23.5} & \textbf{24.5} & 24.9
    & 10.7 & 17.4 & 29.3 & 16.0
    & \textbf{16.8} & \textbf{22.5} & \textbf{24.8} & \textbf{32.2}
    \\
    \bottomrule
    \end{tabular}
    }
    \caption{Results on 11 full-granularity languages.}
    \label{tab:11fulllangs}
    \vspace{-15pt}
    \end{table*}
    

\paragraph{Training.}
We use \instruction{} for two-stage training on Qwen3-8B and evaluate on \benchmark{}.
\textbf{Stage 1: Supervised Fine-Tuning (SFT).}
Using LlamaFactory\footnote{\url{https://github.com/hiyouga/LLaMA-Factory}}, we run full-parameter SFT for five epochs with a cosine LR schedule (peak $10^{-5}$, 10\% warmup), BF16, and DeepSpeed ZeRO-3. The max input length is 32{,}768 tokens. A per-device batch size of 1 with grad-accum 2 yields a global batch size of 16. We validate on \benchmark{} every 500 steps and obtain \codersft{} in about 10 hours.
\textbf{Stage 2: GRPO Reinforcement Learning.}
Starting from \codersft{}, we use verl\footnote{\url{https://github.com/volcengine/verl}} with \textbf{GRPO}, rewarding the length-normalized edit similarity $S$. We train for 15 epochs on roughly 5K tasks (a subset of \instruction{}), with a global batch size of 256 (PPO mini-batch 64; micro-batch 2/GPU), Actor LR $10^{-6}$, and KL penalty 0.001. The max prompt/response lengths are 28{,}672/8{,}192 tokens. This stage performs about 300 gradient updates over 90+ hours, producing \coderl{}.

\paragraph{Model Evaluation.}  We evaluate 30 models in total, including \codersft{} and \coderl{}, using the full evaluation across all languages and granularities. \autoref{tab:7partiallangs} and \autoref{tab:11fulllangs} report the results. These results form the basis of the comparisons and analyses discussed in \autoref{sec:experiment} and \autoref{sec:analysis}.

\subsection{Data Analysis} 

\paragraph{Task Count for Each Language.} As shown in \autoref{fig:taskcount}, the training set $\mathcal{D}_{\text{t}}$ is much larger than the evaluation set $\mathcal{D}_{\text{e}}$, approximately 17K versus 1,286 tasks, giving broad coverage in training while keeping test annotation manageable. Full-granularity languages such as Python and Java receive substantial Class- and Function-level supervision. In contrast, languages like HTML are concentrated at the Block and Line levels, matching their typical usage. In $\mathcal{D}_{\text{e}}$, these patterns persist but are much sparser, especially for Verilog and R at the Class and Function levels, making these slices of the benchmark both rare and highly informative.

\paragraph{Input \& Output Token Distribution.} \autoref{fig:tasktoken} shows a clear context–target imbalance: on average, inputs are more than ten times longer than outputs. C has the heaviest contextual load, with average inputs above 6,000 tokens, around ten times those of Verilog at about 600 tokens. Yet Verilog requires the longest completions, with average outputs around 550 tokens, roughly 2.2 times those of C at about 250 tokens, revealing substantial cross-language variation in token budgets.

\section{Experiment}\label{sec:experiment}
\subsection{Experiment Setup}
\noindent\textbf{Models and Datasets.}
We fine-tune \textbf{Qwen3-8B} with a two-stage pipeline. Training uses \instruction{} and evaluation uses the human-annotated \benchmark{}. All experiments run with \textbf{8$\times$NVIDIA A100-80GB}.

\noindent\textbf{Evaluation Baselines.}
Our evaluation includes general-purpose models such as gpt-4o, o3-mini, and o4-mini~\citep{gpt4,o3ando4}; Claude-3-7-Sonnet and Claude-4-Sonnet~\citep{claude37sonnet,claude4sonnet}; and Gemini-2.5 Pro and Flash~\citep{gemini}. We also assess the Qwen3 series~\citep{qwen3technicalreport,qwq-32b} and the DeepSeek family, along with their distilled variants~\citep{deepseek_r1,deepseek_v3,llama}.

\subsection{Evaluation Metric}
We evaluate LLMs with a \textbf{Length-Normalized Edit Similarity} $S$ defined in \autoref{sec:TaskDefinition}. Raw edit distance (ED) measures disagreement and is therefore inversely related to quality, which makes scores hard to compare across examples of different lengths. We instead convert ED into a similarity ratio $S \in [0,1]$ by normalizing against the longer sequence. This follows standard practice for Levenshtein-based similarity and yields a more interpretable, length-robust metric.

\subsection{Main Result}
Closed-source models such as Claude and Gemini still lead, but strong open-source systems, including Qwen3-Coder-480B-A35B-Instruct and DeepSeek-R1, are closing the gap, particularly on Line and Block tasks. The results show a transparent difficulty gradient: Line is the easiest, Block and Function are in the middle, and Class remains the hardest. Qwen3-Coder-480B-A35B-Instruct maintains stable performance across both full-granularity languages, such as Java and Python, and partial-granularity languages, such as C++ and Rust. At the same time, weaker models fluctuate sharply and depend heavily on the coverage of pretraining data.

\section{Analysis}\label{sec:analysis}

\begin{figure}[t]
    \centering
    \includegraphics[width=0.8\linewidth]{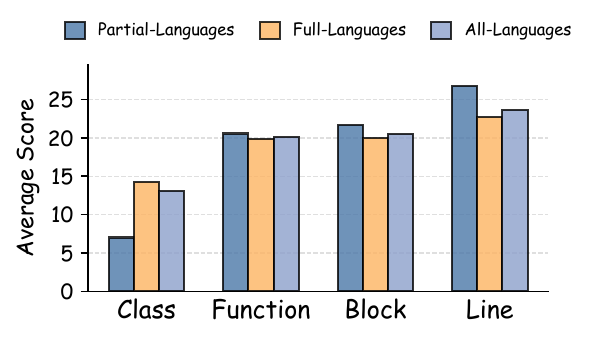}
    \caption{Granularity difficulty in partial-granularity, full-granularity, and all languages.}
    \label{fig:vsbaseline}
\end{figure}

\paragraph{Comparative Analysis.}
\autoref{fig:vsbaseline} shows two main trends. First, model performance consistently drops as we move from Line to Block/Function to Class, confirming that Class-level tasks are the most challenging. Second, full-granularity languages outperform partial-granularity ones at all levels, and this gap grows with task difficulty: smallest at the Line level, largest at the Class level. This suggests that partial-granularity languages are limited by both weaker syntactic coverage and the difficulty of generating long, structured code.

\begin{figure}[t]
    \centering
    \includegraphics[width=1\linewidth]{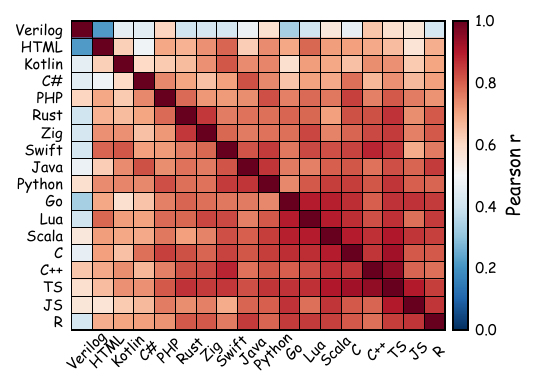}
    \caption{ Pearson correlation of model scores across \languagenums{} languages.}
    \label{fig:heatmap}
\end{figure}

\begin{figure}[t]
    \centering
    \includegraphics[width=1\linewidth]{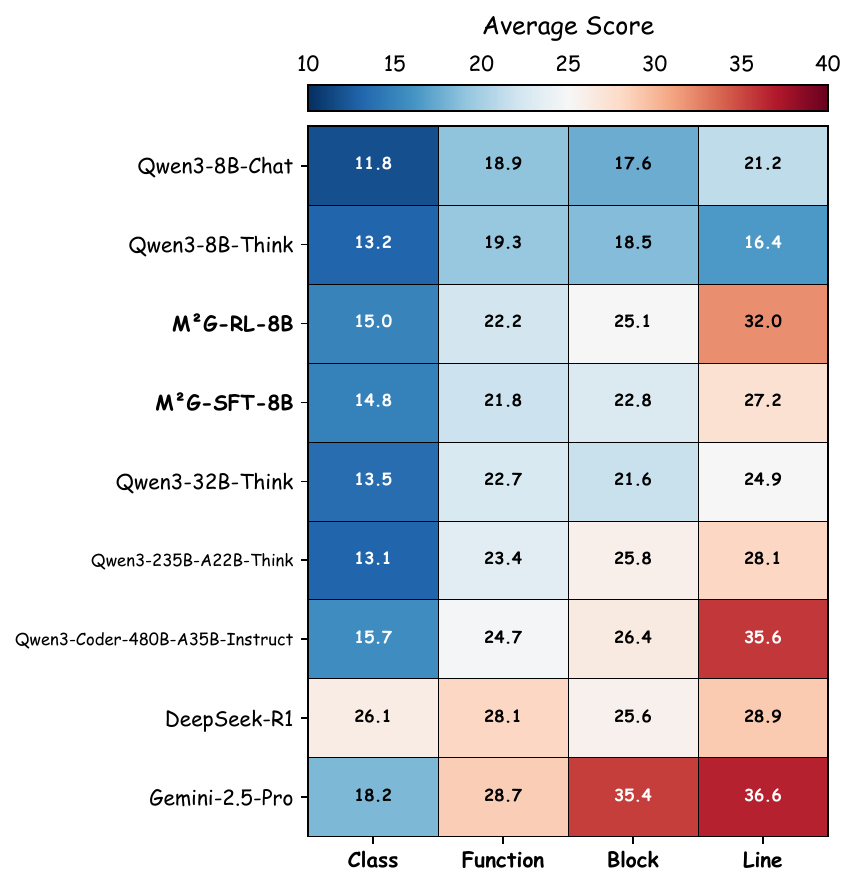}
    \caption{A comparison of the model trained using our method, the base model, and some strong models.} 
    \label{fig:heatmapwithstrongmodel}
    \vspace{-5pt}
\end{figure}

\begin{figure*}[htp]
    \centering
    \includegraphics[width=1\linewidth]{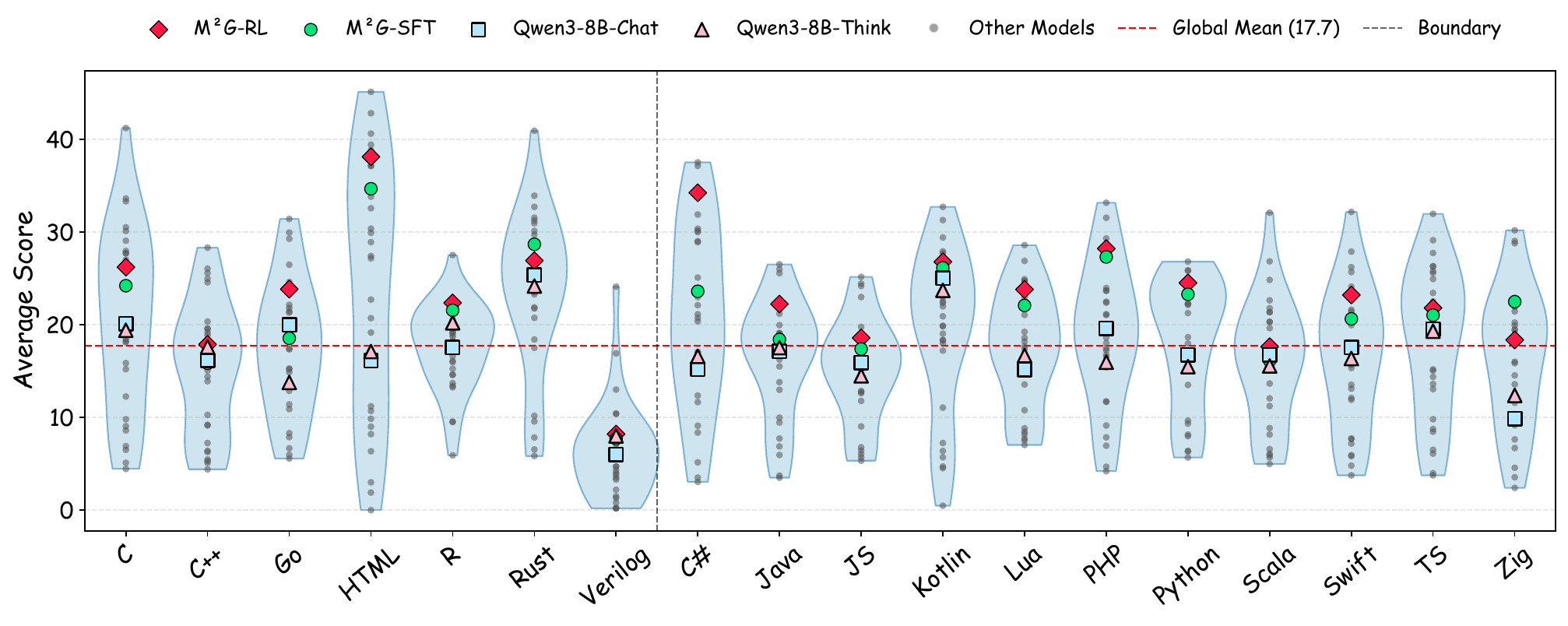}
    \caption{Full-granularity average scores of all models for each language. Violin widths indicate the density of models in a given score range. The orange rhombuses and green circles represent our model; the light blue squares and pink triangles represent the base model; the gray dots represent other models; the horizontal red dashed lines represent the global average; and the vertical gray dashed lines represent the language group boundary.}
    \label{fig:boxplotwithavg}
    \vspace{-15pt}
\end{figure*}

\paragraph{Language Correlation.}
\autoref{fig:heatmap} reports Pearson correlations of model scores across \languagenums{} languages. Most cells are dark red, indicating strong positive correlations for almost all language pairs. This pattern suggests that models learn shared programming concepts rather than memorizing language-specific syntax. We also observe mild clustering by paradigm: for example, Java, C\#, and C++ correlate more strongly with each other than with languages that differ in style and domain, such as Verilog and Kotlin.

\paragraph{Model Quality Comparison.}
In \autoref{fig:heatmapwithstrongmodel}, we compare our models with seven representative baselines. Both the SFT and RL variants clearly outperform the Qwen3-8B base model, while the RL model further closes the gap to Qwen3-235B-A22B-Think and the specialized Qwen3-Coder-480B-A35B-Instruct, despite using only 8B parameters. \autoref{fig:boxplotwithavg} aggregates scores by language and shows that our models consistently lie above the global mean, with the RL model concentrated in the high-score region. This indicates that the proposed training pipeline yields stable, language-agnostic gains.

\section{Related Works}
\paragraph{Code Large Language Models.}
Leveraging advancements in NLP, pretraining techniques have significantly bolstered code understanding and synthesis. Early encoder-based models like CodeBERT~\citep{CodeBERT} and encoder-decoder models like CodeT5~\citep{CodeT5} adopted NLP-inspired architectures and objectives for tasks such as code generation, infilling, summarization, refinement, and translation~\citep{CodeXGLUE,CodeTransOcean,refine_gpt,chat_unitest}. The emergence of code-specific large language models (LLMs)~\citep{starcoder,code_llama,guo2024deepseekcoder,codearena,execrepobench,adc,vgamegym}, exemplified by CodeGen~\citep{codegen} and Code Llama~\citep{code_llama}, demonstrates foundational competence in code understanding and generation. To enhance instruction-following capabilities, recent work has focused on instruction tuning~\citep{instructGPT,llama_adapter,self_instructions}, with innovations such as code Evol-Instruct~\citep{wizardcoder} and the use of real-world code in OSS-Instruct~\citep{magicoder} and CodeOcean~\citep{wavecoder} to improve instruction data quality and realism. Inspired by multi-agent collaboration~\citep{multi_agents_survey,autonomous_agents_survey}, language-specific agents have been introduced to create multilingual instruction datasets, with multilingual benchmarks~\citep{multiple,mceval,mdeval,fullstack,bigcodebench,livereporeflection} assessing these models' cross-lingual capabilities.

\paragraph{Multi-granularity Code Generation.}
While existing code generation benchmarks have made significant progress, they predominantly focus on single-level evaluation. Function-level code generation benchmarks like HumanEval~\citep{humaneval} and MBPP~\citep{mbpp} evaluate standalone function generation, while repository-level benchmarks such as CrossCodeEval~\citep{crosscodeeval} and M2RC-Eval~\citep{m2rceval} assess cross-file generation but treat all tasks uniformly without distinguishing generation contexts. Similarly, code editing benchmarks like CodeEditorBench~\citep{codeeditorbench} and CanItEdit~\citep{editevaluatingabilitylarge} evaluate modification capabilities but typically focus on function-level or single-file edits. This level-agnostic evaluation overlooks the fact that code generation and editing tasks vary substantially across different scopes; completing a single line requires a different context and reasoning than implementing an entire class. Our work addresses this gap by introducing \benchmark{}, a multi-granularity benchmark that systematically evaluates models across four distinct code scopes (class, function, block, and line) in \languagenums{} languages. This design enables fine-grained analysis of model capabilities at each level and provides more comprehensive insights into their strengths and limitations across diverse generation contexts.

\section{Conclusion}
This paper introduces \benchmark{}, a multi-granularity, multilingual evaluation framework assessing LLMs at four granularities (Class, Function, Block, Line). We constructed training and test datasets, trained our \benchmark{}-Coder models using SFT and RL, and evaluated them against 28 other LLMs. The results showed a clear difficulty gradient (Line-level easiest, Class-level hardest) and a performance gap between full- and partial-granularity languages. Nevertheless, the strong cross-language correlation indicates that models learn transferable programming logic. \benchmark{} thus offers a granular approach to measuring code-LLM capabilities, highlighting challenges in complex code generation and in partial-granularity language support.

\section{Limitations}
\benchmark{} has several limitations: (1) imbalanced language coverage, with partial-granularity languages lacking certain task granularities; (2) evaluation focuses on syntactic similarity rather than execution-based correctness; (3) relatively small dataset scale (17K+ training, 1,286 test instances); (4) human annotation, while ensuring quality, limits scalability and may introduce bias.

\section*{Ethics Statement}
All code is collected from public GitHub repositories with permissive licenses. We exclude repositories containing sensitive information and respect original permits. Our evaluation framework may reflect biases in open-source communities. Models trained on this data may inherit these biases. This work is intended for research purposes only and should not replace human judgment in critical applications.

\bibliography{custom} 

\appendix
\onecolumn
\newpage
\section{Prompts for Generation of Code Description and Candidate Inference }\label{prompt}

We present both the system prompt and the user prompt.
\begin{figure}[ht]
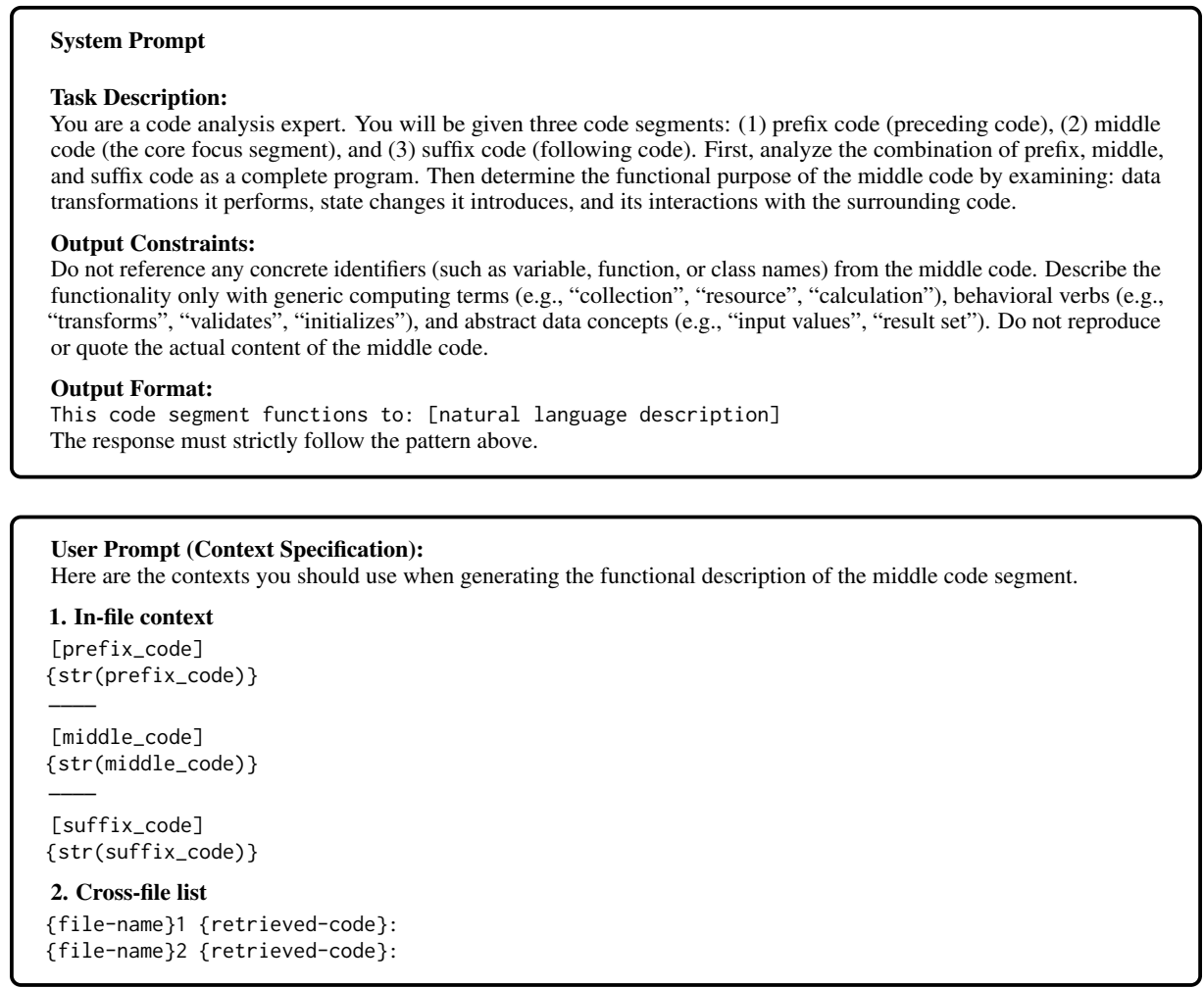

    \centering
    \footnotesize

    \begin{tcolorbox}[colback=white,colframe=black]
    \textbf{System Prompt} \\[2pt]

    \textbf{Task Description:} \\
    You are a code analysis expert. You will be given three code segments:
    (1) prefix code (preceding code),
    (2) middle code (the core focus segment),
    and (3) suffix code (following code).
    First, analyze the combination of prefix, middle, and suffix code as a complete program. 
    Then determine the functional purpose of the middle code by examining:
    data transformations it performs, state changes it introduces, and its interactions with the surrounding code.

    \medskip

    \textbf{Output Constraints:} \\
    Do not reference any concrete identifiers (such as variable, function, or class names) from the middle code. 
    Describe the functionality only with
    generic computing terms (e.g., ``collection'', ``resource'', ``calculation''), 
    behavioral verbs (e.g., ``transforms'', ``validates'', ``initializes''), 
    and abstract data concepts (e.g., ``input values'', ``result set''). 
    Do not reproduce or quote the actual content of the middle code.

    \medskip

    \textbf{Output Format:} \\
    \texttt{This code segment functions to: [natural language description]} \\
    The response must strictly follow the pattern above.
    \end{tcolorbox}

    \vspace{4pt} 

    \begin{tcolorbox}[colback=white,colframe=black]
    \textbf{User Prompt (Context Specification):} \\
    Here are the contexts you should use when generating the functional description of the middle code segment.

    \medskip
    \textbf{1. In-file context} \\[2pt]
    \texttt{[prefix\_code]} \\
    \texttt{\{str(prefix\_code)\}}

    \vspace{2pt}

    \texttt{------------} \\[2pt]
    \texttt{[middle\_code]} \\
    \texttt{\{str(middle\_code)\}}

    \vspace{2pt}

    \texttt{------------} \\[2pt]
    \texttt{[suffix\_code]} \\
    \texttt{\{str(suffix\_code)\}}

    \medskip
    \textbf{2. Cross-file list} \\[2pt]
    \texttt{\{file-name\}1 \{retrieved-code\}:} \\
    \texttt{\{file-name\}2 \{retrieved-code\}:}
    \end{tcolorbox}
    \vspace{-4mm}
    \caption{System (top) and user (bottom) prompts for generating abstract functional descriptions of code segments.}
    \label{fig:prompt_code_both}
\end{figure}

\begin{center}
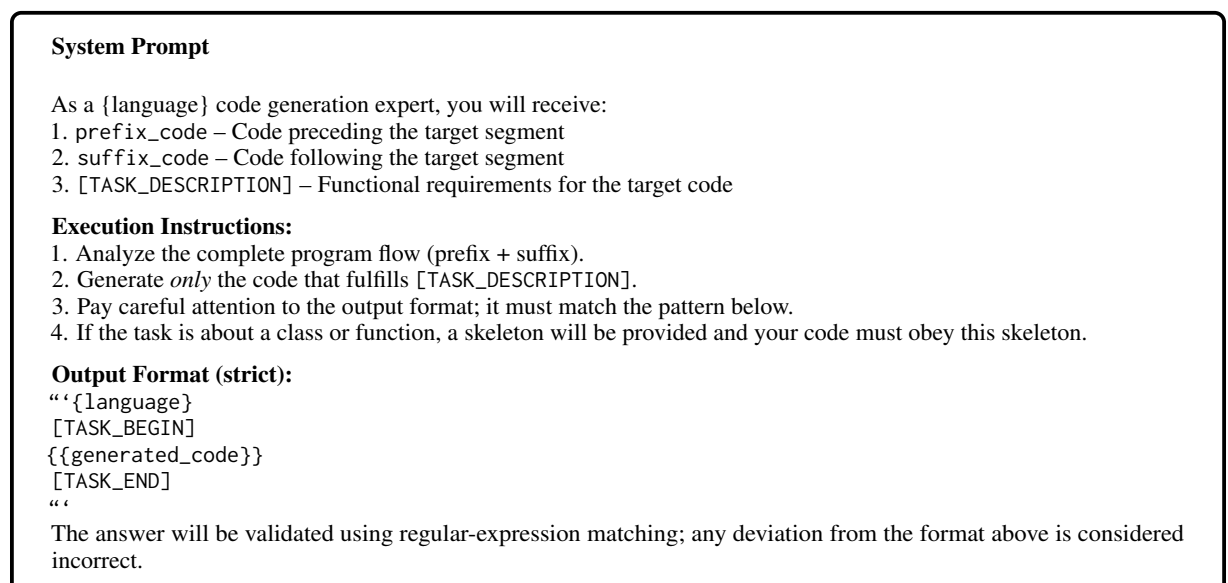

\footnotesize
\begin{tcolorbox}[colback=white,colframe=black]
\textbf{System Prompt} \\[2pt]

As a \{language\} code generation expert, you will receive: \\
1. \texttt{prefix\_code} -- Code preceding the target segment \\
2. \texttt{suffix\_code} -- Code following the target segment \\
3. \texttt{[TASK\_DESCRIPTION]} -- Functional requirements for the target code

\medskip

\textbf{Execution Instructions:} \\
1. Analyze the complete program flow (prefix + suffix). \\
2. Generate \emph{only} the code that fulfills \texttt{[TASK\_DESCRIPTION]}. \\
3. Pay careful attention to the output format; it must match the pattern below. \\
4. If the task is about a class or function, a skeleton will be provided and your code must obey this skeleton.

\medskip

\textbf{Output Format (strict):}
\begin{flushleft}
\texttt{```\{language\}} \\
\texttt{[TASK\_BEGIN]} \\
\texttt{\{\{generated\_code\}\}} \\
\texttt{[TASK\_END]} \\
\texttt{```}
\end{flushleft}
The answer will be validated using regular-expression matching; any deviation from the format above is considered incorrect.
\end{tcolorbox}
    \vspace{-4mm}
\captionof{figure}{System prompt for code generation with strict output formatting.}
\label{fig:prompt_codegen_system}
\end{center}
\begin{center}
\footnotesize
\begin{tcolorbox}[colback=white,colframe=black]
\textbf{User Prompt (Context Specification):} \\
Here are the contexts you can use when generating the target code.

\medskip

\textbf{\{task\_type\}}

\medskip
\textbf{Current File:}
\begin{flushleft}
\texttt{```\{language\}} \\
\texttt{\{prefix\_code\}} \\
\texttt{[TASK\_START]} \\
\texttt{[TASK\_DESCRIPTION \{code\_description\}]} \\
\texttt{[SKELETON \{skeleton\}]} \\
\texttt{[TASK\_END]} \\
\texttt{\{suffix\_code\}} \\
\texttt{```}
\end{flushleft}

\medskip

\textbf{Cross-file list:}
\begin{flushleft}
\texttt{\{file-name\}1 \{retrieved-code\}:} \\
\texttt{\{file-name\}2 \{retrieved-code\}:}
\end{flushleft}
\end{tcolorbox}
\vspace{-4mm}

\captionof{figure}{User prompt specifying in-file and cross-file contexts for code generation.}
\label{fig:prompt_codegen_user}
\end{center}

\noindent\textbf{Usage Summary.}
This appendix lists the exact natural-language prompts used in our pipeline.
\autoref{fig:prompt_code_both} provides the prompts for generating abstract functional descriptions of code segments, which are used to construct the textual descriptions $d$ in \instruction{}.
\autoref{fig:prompt_codegen_system} and \autoref{fig:prompt_codegen_user} show the prompts for multi-granularity code generation, which are used both to filter training tasks and to query models during evaluation on \benchmark{}.

\end{document}